\documentclass{article}
\usepackage{collas2022_conference,times}
\usepackage{amsmath,amsfonts,bm}

\def\figref#1{figure~\ref{#1}}
\def\eqref#1{equation~\ref{#1}}
\def\1{\bm{1}}

\def\va{{\bm{a}}}

\def\vp{{\bm{p}}}

\def\vr{{\bm{r}}}
\def\vs{{\bm{s}}}

\def\vx{{\bm{x}}}

\def\eva{{a}}

\def\mW{{\bm{W}}}

\DeclareMathAlphabet{\mathsfit}{\encodingdefault}{\sfdefault}{m}{sl}
\SetMathAlphabet{\mathsfit}{bold}{\encodingdefault}{\sfdefault}{bx}{n}

\def\sA{{\mathbb{A}}}

\usepackage{mathtools}
\usepackage{amssymb}
\usepackage[stretch=10]{microtype}
\usepackage{acronym}
\usepackage{xcolor}
\usepackage{colortbl}
\usepackage{booktabs}
\usepackage{array}
\usepackage{multirow}

\usepackage{graphicx,amsmath,amssymb,subcaption,url,xspace}
\usepackage{anyfontsize}
\newcommand{\eg}{e.g.,\xspace}

\newcommand{\etc}{etc.\@\xspace}

\usepackage[symbol]{footmisc}

\usepackage{hyperref}
\usepackage{cleveref}
\hypersetup{
    colorlinks=true,
    linkcolor=red,
    filecolor=magenta,
    urlcolor=blue,
    citecolor=purple,
    pdftitle={Overleaf Example},
    pdfpagemode=FullScreen,
    }

\acrodef{WADE}[WADE]{Weighted Average Data Efficiency}
\acrodef{RNN}[RNN]{recurrent neural network}

\newlength{\currentparskip}

\def\smtt#1{\texttt{\textsc{#1}}}

\title{Benchmarking learning efficiency \\ in deep reservoir computing}

\author{Hugo Cisneros \\
   CIIRC\footnotemark[1]\ , WILLOW\footnotemark[2]\\
\texttt{hugo.cisneros@cvut.cz} \\
\And % Use And to have authors side by side
Josef Sivic \\
CIIRC \\
\texttt{josef.sivic@cvut.cz}\\
\And % Use AND to have authors block one under the other
Tomas Mikolov \\
CIIRC \\
\texttt{tomas.mikolov@cvut.cz}\\
\\
}

\collasfinalcopy % Uncomment for camera-ready version, but NOT for submission.

\begin{document}

\maketitle

\begin{abstract}
  It is common to evaluate the performance of a machine learning model by
  measuring its predictive power on a test dataset. This approach favors
  complicated models that can smoothly fit complex functions and generalize well
  from training data points. Although essential components of intelligence,
  speed and data efficiency of this learning process are rarely reported or
  compared between different candidate models. In this paper, we introduce a
  benchmark of increasingly difficult tasks together with a data efficiency
  metric to measure how quickly machine learning models learn from training
  data. We compare the learning speed of some established sequential supervised
  models, such as RNNs, LSTMs, or Transformers, with relatively less known
  alternative models based on reservoir computing. The proposed tasks require a
  wide range of computational primitives, such as memory or the ability to
  compute Boolean functions, to be effectively solved. Surprisingly, we observe
  that reservoir computing systems that rely on dynamically evolving feature
  maps learn faster than fully supervised methods trained with stochastic
  gradient optimization while achieving comparable accuracy scores. The code,
  benchmark, trained models, and results to reproduce our experiments are
  available at {\small\url{https://github.com/hugcis/benchmark_learning_efficiency/}}.
\end{abstract}

\renewcommand{\thefootnote}{\fnsymbol{footnote}}
   \footnotetext[1]{Czech Institute of Informatics, Robotics and Cybernetics,
     Czech Technical University in Prague.}
   \footnotetext[2]{WILLOW project, Inria and Département d’Informatique de l’École Normale Supérieure, PSL Research University.}
\renewcommand{\thefootnote}{\arabic{footnote}}

\section{Introduction}
Most machine learning models are evaluated by measuring performance on a
specific dataset or task. Learning efficiency -- the ability to learn,
generalize, and adapt quickly from a few examples -- is crucial for practical
intelligence \citep{kanazawaGeneralIntelligenceDomainspecific2004} as well as
low-data machine learning applications yet rarely used to evaluate models.
Supervised learning systems are theoretically limited in their learning speed by
the optimization algorithms used for training. These algorithms such as
stochastic gradient descent (SGD) have various speed guarantees depending on the
structure of the function to be
optimized~\citep{bottouOptimizationMethodsLargescale2018}. However, when
intelligent beings learn, they appear to quickly re-use past knowledge and
progressively improve over time. Their learning speed depends on a dynamically
evolving internal state.
To measure the learning efficiency of various systems, we propose in this work the \ac{WADE}
metric based on the time taken to reach several test accuracy checkpoints. We
also design a simple modular benchmark composed of a set of sequential tasks.
They begin with the task of recognizing a simple periodic sequence in an input
string and end with elaborate question answering tasks that require counting
occurrences of patterns and long-term memory.

Established sequential supervised models such as recurrent neural networks
\cite[RNNs; ][]{elmanFindingStructureTime1990}, long short-term memory networks
\cite[LSTMs;][]{hochreiterLongShortTermMemory1997} or Transformers
\citep{vaswaniAttentionAllYou2017} lack essential properties such as learning
beyond the training phase or the ability to adapt over time after being trained. These models can also be expensive to train, requiring a large number
of labeled training examples to reach reasonable performance, leading to poor
learning speeds. In this work, we use the newly proposed \ac{WADE} metric and the benchmark dataset to experimentally compare the learning speed of these well established models, such as RNNs, LSTMs and Transformers, to less explored {\em reservoir computing} models.

Reservoir computing is a computational framework that aims to exploit the states
of a complex dynamical system. The simplest example of a reservoir computer is a
\ac{RNN} with frozen weights. This special \ac{RNN} performs random manipulation
on its hidden state in reaction to each new input. 
Interestingly, it has been shown that with a
specific initialization of the frozen weights, these \ac{RNN}s (called Echo
state networks) can keep a memory of past inputs
\citep{jaegerEchoStateApproach2001}. 
Usually, a standard linear regression is
added as a decoder to extract valuable representation from the hidden state for
some downstream task. Freezing the weights of these recurrent models is useful
when available supervision is very limited or non-existent or for reinforcement
learning with sparse rewards since direct training would be impossible. In such
cases a reservoir computer creates a continuously evolving pool of random
functions that can be combined using the last trainable layer.

When evolving in response to input stimuli, complex recurrent systems such as
\acp{RNN} are building dynamically changing representations of data within their
internal state \citep{boccaraModelingComplexSystems2010}. We know that these
internal states can be interesting on their own because of their ability to
self-organize and exhibit increasingly complex behaviors
\citep{koppelAlmostMachineindependentTheory1991,
  bennettLogicalDepthPhysical1995, allenEvolutionEmergenceLearning2003,
  goldsteinEmergenceComplexSystems2011, cisnerosEvolvingStructuresComplex2019}.
In this paper we wish to investigate whether complex dynamical systems --- in particular RNNs with frozen weights (echo-state networks) \citep{jaegerEchoStateApproach2001} and reservoir cellular automata \citep{yilmazReservoirComputingUsing2014} ---  
create representations that allow them to learn faster as measured by our
metric.
\vspace{-8pt}
\paragraph{Contributions.} In this paper, we make the following main contributions: First, we introduce the
\acf{WADE} metric to measure the learning speed of various learning systems and use it to benchmark a few standard models on the IMDB text
classification task \citep{maasLearningWordVectors2011}.
Second, we present a benchmark of language-based tasks of increasing difficulty
to evaluate the learning speed in different conditions. The proposed tasks require a wide range of computational primitives, such as memory or the ability to compute Boolean functions, to be effectively solved.  Third, we study the
learning speed of reservoir computing learning models 
and compare them with more standard supervised solutions.
\vspace{-8pt}
\section{Related work}
The \ac{WADE} metric is a generalization of the \emph{Time-to-threshold} metric
\citep{taylorCrossdomainTransferReinforcement2007,
  taylorTransferLearningInterTask2007} introduced for measuring transfer
learning in reinforcement learning contexts. In general, the Time-to-threshold
is simply defined as the number of training steps needed to reach a fixed
threshold performance. However this definition leaves open the choice of
threshold or the definition of a training step. \ac{WADE} alleviates this issue
by aggregating several of these thresholds into a single number that summarizes
the learning speed.

Other metrics for measuring how quickly a model adpats to new tasks have been
introduced in the context of transfer learning, few-shot and zero-shot learning.
In few-shot learning, one tries to obtain the best performance for a particular
task using a small amount of labeled data compared to the task's fully
supervised equivalent~\citep{wangGeneralizingFewExamples2020}. This correlates
with a model's learning speed, but these problems often measure how much prior
information about similar data has been encoded in the models. With our
benchmark and the \ac{WADE} metric, we explicitly measure the number of steps to reach
multiple test accuracy values using all the data needed, effectively emphasizing
data efficiency.

Sample efficiency has also been studied in the context of
reinforcement learning. \cite{chevalier-boisvertBabyaiPlatformStudy2018} use the
number of demonstrations before a task is solved to measure sample efficiency.
This requires defining what solving the task means, which may vary from task to
task. Another approach is to measure performance (cumulated reward, accuracy,
\etc) after a fixed budget of training steps
\citep{yaratsImprovingSampleEfficiency2019}. In this case, the most efficient
model is the one that achieves the best performance within the allocated budget.
In other cases, the sample efficiency is mentioned but not explicitly measured
and one has to examine the learning curves
\citep{buckmanSampleefficientReinforcementLearning2018}. The \ac{WADE} metric is
a general approach to measure the learning efficiency of machine learning
models. We use it to benchmark a few standard models on the IMDB text
classification tasks \citep{maasLearningWordVectors2011} and propose a set of
modular and extensible language based tasks.

Synthetic tasks such as ours have played a vital role in a series of crucial
advances in machine learning algorithms. For example, the XOR problem has
partially motivated the development of neural networks
\citep{minskyPerceptronsIntroductionComputational1972,
  rumelhartLearningInternalRepresentations1985}, and the \emph{circle and ring}
dataset has inspired the creation of novel clustering algorithms
\citep{ngSpectralClusteringAnalysis2001}. The design of synthetic datasets has also been 
an essential component of the development of learning algorithms with
memory and general computational capabilities
\citep{hochreiterLongShortTermMemory1997,
  joulinInferringAlgorithmicPatterns2015, gravesNeuralTuringMachines2014,
  westonAICompleteQuestionAnswering2016, richardsonProbingNaturalLanguage2020}.

Other tasks are based on real datasets with artificial manipulations
\citep{krizhevskyLearningMultipleLayers2009, srivastavaCompeteCompute2013a,
  goodfellowEmpiricalInvestigationCatastrophic2014,
  nguyenVariationalContinualLearning2017}. The goal of our dataset is to be
truly progressive in difficulty yet simple to understand and extend, to allow
applications in the field of online learning, and to easily understand a model's
basic computational capacities. Combined with our metric, it enables us to
measure learning speed across a range of conditions. In contrast to similar
synthetic datasets, we built this benchmark so that the last task is vastly more
complicated than the first and could still be extended to more complex examples.

\section{A benchmark for reservoir computing\label{sec:tasks}}
To measure the learning speed of candidate systems and their ability to improve
over time, we propose a performance metric and a standardized set of tasks. We
want to select those systems that quickly and reliably adapt and learn from new inputs.
For this purpose, we introduce the Weighted Average Data Efficiency (WADE)
metric. It aggregates the speed at which a model reaches several test accuracy
checkpoints. We describe the metric in more detail in Section
\ref{sec:performance-metric}.

To reliably compare learning speeds for various systems on a shared foundation,
we also introduce a novel dataset described in Table~\ref{tab:all-tasks}. It is
made up of sequential tasks that begin with straightforward pattern recognition
and progressively increase in complexity to approach the complexity of natural
language and other complex real-world tasks.

We do not focus on the prediction performance of our models, but rather on their
data efficiency --- the number of example sequences they need to learn from before
reaching a target accuracy on a validation set.

\begin{table}[htbp]
  \centering
{\fontsize{8}{8.0}\selectfont
  \begin{tabular}{cp{.37\linewidth}p{.48\linewidth}}
    \toprule
    \bfseries Task id & \bfseries Name & \bfseries Description \\
    \midrule
    1 & Simple periodic pattern identification & Identify a simple periodic pattern. \\
    \arrayrulecolor{black!20}\specialrule{0.2pt}{.2em}{.4em}
    2 & Harder periodic pattern identification & Identify a periodic pattern with an arithmetically
                            increasing period. \\
    \arrayrulecolor{black}\midrule
    3 & Symbol counting & Count symbols from a sequence. \\
    \arrayrulecolor{black!20}\specialrule{0.2pt}{.2em}{.4em}
    4 & Pattern counting & Count patterns (delimited group of symbols) from a sequence. \\
    \arrayrulecolor{black}\midrule
    5 & Simple question answering & Answer simple YES/NO questions from a single prompt. \\
    \arrayrulecolor{black!20}\specialrule{0.2pt}{.2em}{.4em}
    6 & Harder question answering & Answer simple YES/NO questions from a single prompt
                                    with a more extensive vocabulary. \\
    \arrayrulecolor{black!20}\specialrule{0.2pt}{.2em}{.4em}
    7 & Question answering with world definition & Answer YES/NO questions from a
                                                   sequence of prompts. \\
    \arrayrulecolor{black!20}\specialrule{0.2pt}{.2em}{.4em}
    8 & Question answering with world definition and counting & Answer YES/NO
                                                                and counting questions from a
                                                                sequence of prompts. \\
    \arrayrulecolor{black!20}\specialrule{0.2pt}{.2em}{.4em}
    9 & Adjective question answering & Answer YES/NO and adjective questions from a
                                       sequence of prompts. \\
    \arrayrulecolor{black!20}\specialrule{0.2pt}{.2em}{.4em}
    10 & Adjective question answering and counting & Answer YES/NO, adjective,
                                                     and counting
                                                     questions from a
                                                     sequence of prompts. \\
    \arrayrulecolor{black}\bottomrule
  \end{tabular}
}\vspace{-5pt}
  \caption{General description of all the tasks in the benchmark.}
  \label{tab:all-tasks}
\end{table}
Standard benchmarks and metrics such as those introduced in this paper have
always been essential in advancing various aspects of machine learning. For
example, the LSTM network demonstrated a superior memory capacity on a set of
synthetic tasks designed to challenge the memory of sequential learning systems
\citep{hochreiterLongShortTermMemory1997}. Our goal with this benchmark is to
emphasize measuring learning speed across tasks of varying difficulties with a
range of computational requirements rather than focusing on performance only. We describe the performance metric and the benchmark next.

\subsection{Performance metric\label{sec:performance-metric}}

We introduce the Weighted Average Data Efficiency (WADE) metric as a way to
measure how quickly a model learns, using a weighted average of inverse times
taken to reach various test accuracy \emph{checkpoints} over time.

\setlength{\currentparskip}{\parskip}
\begin{minipage}[htbp]{.65\linewidth}
   \setlength{\parskip}{\currentparskip}% restore the value

It is computed for an evenly distributed set of target accuracies $\sA$. They
represent the \emph{checkpoints} at which the speed of learning is estimated.
For example, we may choose $\sA = [0.1, 0.2, 0.3, 0.4, \ldots, 1.]$. The metric is
then calculated as 
\begin{equation}
\text{WADE}(\va) = \frac{1}{\sum \alpha} \sum_{\alpha \in \mathbb{A}}\frac{\alpha}{\text{T}(\alpha, \va)},
\label{eq:wade}
\end{equation}
where $\va = (\eva_{0}, \eva_{1}, \ldots, \eva_{n})$ is a sequence of test accuracies achieved by the evaluated system
sampled at different training steps. The quantity $a_{i}$ typically
corresponds to the accuracy reached after seeing $i$ examples, and
$\text{T}(\alpha, \va)$ is the number of steps in the sequence $\va$ needed to reach
an accuracy of $\alpha$. It is defined as 
\begin{equation}
  \text{T}(\alpha, \va) = \min\left\{ i \in \{1, \ldots, n, + \infty \}\; |\; a_{i} \geq \alpha \right\}.
\label{eq:tto}
\end{equation}

We also define $\text{T}(\alpha, \va) = +\infty$ if the accuracy value $\alpha$ is never
reached in $\va$. This is equivalent to appending an additional term $\eva_{+\infty}$
to $\va$, always set to the maximum accuracy 1. Note that by construction,
$\text{T}(\alpha, \va)$ is in $[1, + \infty [$.
\end{minipage}
\hfill
\begin{minipage}[htbp]{.330\linewidth}
 \includegraphics[width=.9\linewidth]{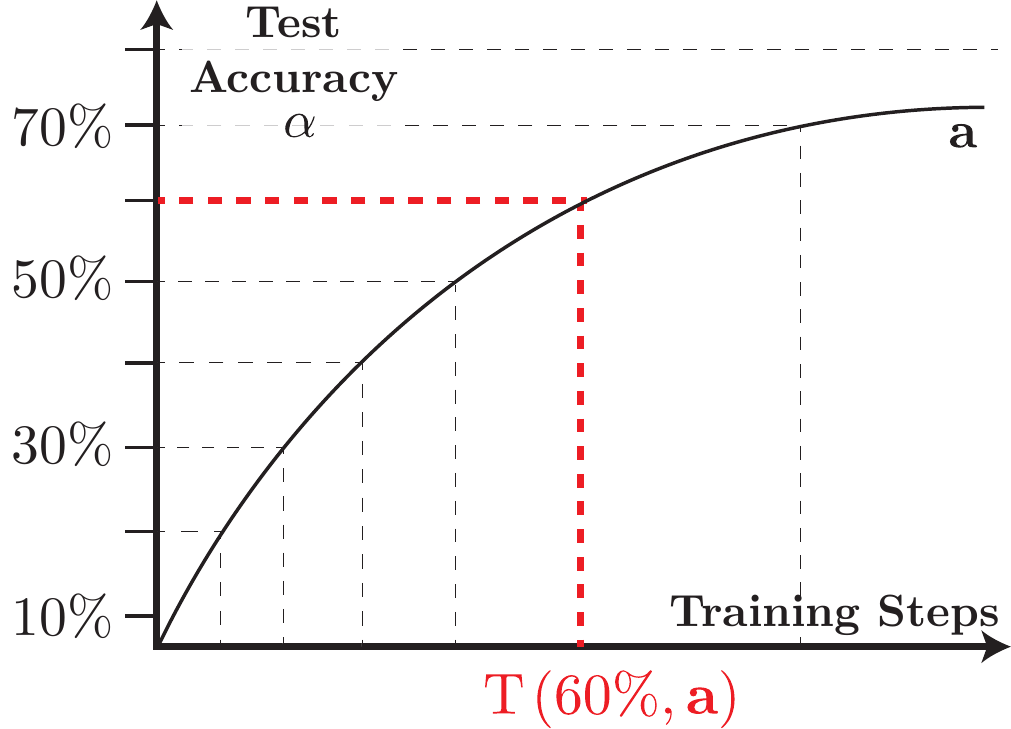}
 \vspace{-7pt}
 \captionof{figure}{Illustration of the calculation of $\text{T}(\cdot, \cdot)$,
   representing the number of training steps (x-axis) needed to reach a certain test
   accuracy $\alpha$ (y-axis) from a learning curve. In this example,
   $\sA = [0.1, \ldots, 0.8]$ (y-axis). $\text{T}(0.6, \va)$ is highlighted in red.
   $\text{T}(0.8, \va) = +\infty$ as the accuracy of 0.8 is never reached.}
  \label{fig:metric_tto}
\end{minipage}

Since $\text{T}$ can be $+\infty$ we define
$\frac{1}{+\infty} = 0$ for the quantity in \eqref{eq:wade} to always exist. A visual
intuition of $\text{T}(\cdot, \cdot)$ is given in \figref{fig:metric_tto}.

The choice of checkpoints $\sA$ does not need to be tuned in any specific way
because $\text{WADE}(\va)$ converges quickly to a single value when $\sA$
approaches the continuous interval $[0, 1]$. The approximation is good enough as
long as $\sA$ is not too coarse (more than ten elements was enough in our
experiments) and the WADE values computed from the same set $\sA$ are
comparable.

The time-to-threshold $\text{T}$ is always above or equal to 1 step for any threshold and sequence of accuracy scores.
We have
$\forall\: \alpha \in [0, 1],\;  \forall\: \va = (\eva_{n})_{I \subset \{\mathbb{N}\; \cup \{+\infty\} \}}$, it holds that
\begin{equation}
  \label{eq:wade-0-1}
\frac{1}{\text{T}(\alpha, \va)} \leq 1,
\end{equation}
and therefore we always get that $0 \leq \text{WADE}(\va) \leq 1$. The metric is equal to $0$
for systems that never get past the smallest possible accuracy while a $1$
corresponds to reaching a perfect test accuracy in only one training step. Such
a system would also be considered to be performing well according to the
underlying performance metric with which it is usually evaluated. Therefore,
maximizing WADE also maximizes performance.

\subsection{Description of tasks in the benchmark\label{sec:descr-tasks-benchm}}

This section provides a more detailed description of each task in our benchmark.
The tasks are designed to be language modeling tasks, where the goal is to
predict some tokens from sequences of previously processed tokens. An overview
of the tasks is given in Table~\ref{tab:all-tasks}. The tasks are divided into three
groups: (i) binary tasks -- with only binary symbols, (ii) general symbolic tasks
-- symbolic manipulations with arbitrary symbols -- and (iii) language-based tasks
-- the symbols represent words in English and behave like a language. We introduce
this benchmark together with the WADE metric, but both can be used in other
contexts as well to measure the learning speed of other systems. Individual
sentences are generated and divided into a training set and a test set for
periodic evaluation of the test accuracy. We give a more detailed description of
each task below\footnote{The tasks are also available as a
  \href{https://github.com/hugcis/incremental_tasks}{Python package on GitHub}.}:

\subsubsection{Binary}
\paragraph{Simple periodic pattern identification.}

The goal of the periodic binary task is for the model to learn a fixed-length
regular pattern. As the system is presented with new binary input tokens, it has
to learn the periodic pattern on the fly and correctly predict the next token. A
pattern of size $n$ is chosen at random and repeated $k$ times to produce a
sequence of length $n \times k$. Examples include: {\fontsize{9}{10.8}\selectfont
  \begingroup \addtolength{\jot}{-.4em}
\begin{align*}
  \begin{split}
  \smtt{\textbf{01}01010101010101010101010101010101} & \quad\text{Pattern with period 2}\\
  \smtt{\textbf{0011}001100110011001100110011001100} & \quad\text{Pattern with period 4}\\
  \texttt{\textbf{011}0110110110110110110110110110110} & \quad\text{Pattern with period 3}
\end{split}\label{eq:1}
\end{align*}
\endgroup
}

\paragraph{Harder periodic pattern identification.}

For this task, we also draw a random binary pattern of size $n$. Each of its
symbols is repeated $k$ times, with $k$ increasing monotonically from 1. A
successful model must learn the pattern on the fly and correctly implement the
arithmetic increase in the size of the period. We set the pattern length to increase by
1 every period in our experiments, but this value can be changed.

\subsubsection{Symbolic counting}

These tasks consist of reading patterns from an input sequence and answering a
simple query about the number of patterns. Unlike the previous tasks, these
require implementing a form of addressable memory that can be queried after the prompt has
ended.

\paragraph{Basic symbol counting.}
The first version of the counting task focuses on counting single symbols from
an input sequence. The sequence ends with a \emph{query} for the count of
one of the symbols. The goal is to predict the last token (in bold) of sequences
of the following form:
{\fontsize{8}{9.6}\selectfont
  \begingroup
\addtolength{\jot}{-.5em}
\begin{align*}
  \underbrace{\smtt{AABBCBABAAB}}_{\text{Input symbols}}
  & \underbrace{\texttt{x}}_{\text{QS}} \texttt{A}
    \underbrace{\textbf{\texttt{5}}}_{\text{Answer}} \\
\end{align*}
\endgroup
}The symbol \texttt{x} is the query symbol (QS) that marks the beginning of the
query. In the first example above, the goal is to predict the token \texttt{5}
because the symbol \texttt{A} appears 5 times. As detailed in
Sect.~\ref{sec:compared-methods}, we represent these nonbinary symbols with
one-hot encoding, so the numerical nature of some tokens is not encoded a priori.

\paragraph{Pattern counting.}
This aim of this task is to count the number of occurrences of delimited patterns
instead of single symbols. A sequence is still divided between a prompt --- before
\texttt{x} --- and a query --- after \texttt{x}. One has to predict the symbol
coming after each separator symbol (S) \texttt{y} in the query part of the
sentence. For example, sentences are of the form: {\fontsize{8}{9.6}\selectfont
\begin{align*}
  &\underbrace{\texttt{AA}}_{\text{Pattern 1}}\underbrace{\texttt{y}}_{\text{S}}
    \underbrace{\texttt{BBC}}_{\text{Pattern 2}}\underbrace{\texttt{y}}_{\text{S}}\underbrace{\texttt{BAB}}_{\text{Pattern 3}}\underbrace{\texttt{y}}_{\text{S}}\underbrace{\texttt{AA}}_{\text{Pattern 4}}\underbrace{\texttt{y}}_{\text{S}}\underbrace{\texttt{B}}_{\text{Pattern 5}}
    \underbrace{\texttt{x}}_{\text{QS}}\underbrace{\texttt{AAy}}_{\text{Query 1}}\underbrace{\texttt{\textbf{2}}}_{\text{Answer 1}}
    \underbrace{\texttt{By}}_{\text{Query 2}}\underbrace{\texttt{\textbf{1}}}_{\text{Answer 2}} \\
\end{align*}}Multiple queries are presented successively, which requires keeping and being
able to retrieve several counts simultaneously. A query is composed of a
pattern, a separator symbol, and the pattern count that the system should
predict.

\subsubsection{Basic language understanding}
To make the tasks progressively more complex, we steer them towards general
language understanding tasks. The tasks described below  are generated automatically, but gradually incorporate more complex skills
required for advanced language processing. 
The last task is a step towards understanding general language albeit with a limited vocabulary. 

\paragraph{Elementary question answering (QA).}
This task introduces elements of natural language. Each example is composed of a
stated fact and a question about that fact. A sentence is constructed from a few
basic elements:
(i) Names (\eg\texttt{JOHN}, \texttt{JAMES}, \etc),
(ii) Verbs (\eg\texttt{HEAR}, \texttt{SEE}, \etc),
(iii) Answers (\texttt{YES} or \texttt{NO}),
(iv) Additional words and symbols (\texttt{I, DO, NOT, AND, BUT, ?, .}).

A random subset of names is selected and we generate a random prompt/question
pair from it.
The question is drawn to ensure an equal proportion of
positive and negative answers. For example, sentences may look like this:
{\fontsize{8}{9.6}\selectfont
  \begingroup
\addtolength{\jot}{-.5em}
\begin{align*}
&  \texttt{I HEAR JOHN AND PAUL .}\texttt{ DO I HEAR PAUL ?} \texttt{\textbf{ YES}}\\
&    \texttt{I SEE JOHN BUT I DO NOT SEE PAUL AND TOM . DO I SEE TOM ? \textbf{NO}}
\end{align*}
\endgroup}
The only token to predict is the binary answer \texttt{YES} or \texttt{NO}.

\paragraph{Question answering (QA) with adjectives.}

This task extends the previous task by adding adjectives and modifiers to the
object names. The queries may be about the subject-verb relation or the
subject-adjective relation.
{\fontsize{8}{9.6}\selectfont
  \begingroup
\addtolength{\jot}{-.5em}
\begin{align*}
  &\texttt{I SEE A SMALL BANANA .}\texttt{ WHAT IS THE SIZE OF THE BANANA I SEE ? } \texttt{\textbf{SMALL}}\\
  &  \texttt{I SEE A LARGE GREEN APPLE BUT I DO NOT SEE A RED APPLE .} \\
  & \quad\quad\texttt{DO I SEE A LARGE APPLE ? \textbf{YES}} \\
  &  \texttt{I SEE A SMALL GREEN APPLE BUT I DO NOT SEE A BANANA .}\\
  &\quad\quad\texttt{WHAT IS THE COLOR OF THE APPLE I SEE ? \textbf{GREEN}}
\end{align*}
\endgroup
}
Here the task output space is slightly larger because the model may be
predicting \texttt{YES}, \texttt{NO}, \texttt{SMALL}, \texttt{GREEN}, \etc

\paragraph{Question answering (QA) with world definition.}

This task introduces more complex configurations in which the state of the world is
defined in one or more sentences, and an unknown number of questions follow.
This is more akin to a real-world conversation where stored facts should be
remembered longer and accessed on demand.
For example, below we show a generated group of sentences followed by several questions:
{\fontsize{8}{9.6}\selectfont
  \begingroup
\addtolength{\jot}{-.5em}
\begin{align*}
  &\texttt{I SEE A SMALL BANANA .}\\
  &  \texttt{I SEE A LARGE GREEN APPLE BUT I DO NOT SEE A RED APPLE .} \\
  &  \texttt{I HEAR A SMALL GREEN APPLE BUT I DO NOT SMELL A BANANA .}\\
  &\quad\quad\texttt{WHAT IS THE COLOR OF THE APPLE I SEE ? \textbf{GREEN}}\\
  &\quad\quad\texttt{HOW MANY THINGS DO I SMELL ? \textbf{ONE}}\\
  & \quad\quad\texttt{DO I SEE A LARGE APPLE ? \textbf{YES}}. \\
\end{align*}
\endgroup}
The difficulty of each of these tasks can be modulated by changing the size of the
base vocabulary, the length of sequences, or the number of queries. Our particular setting of these parameters will be described in section~\ref{sec:task-gen-params}.

\section{Evaluating speed of learning on a standard language classification task}

To show the usefulness of measuring \ac{WADE} on standard language classification tasks, we
train various text classifiers on the IMDB dataset and compare their \ac{WADE}
scores with a usual performance metric for classification: accuracy. This
classification task, first proposed by \cite{maassRealTimeComputingStable2002},
consists of deciding if a movie review is positive or negative from its text. It
contains 25000 training examples and 25000 test examples. The two labels are
balanced in both the training and test set. This dataset is a high dimensional
language-based task with a binary output.

We study five standard models: (i) an Elman recurrent neural network (RNN)
\citep{elmanFindingStructureTime1990} with $\tanh$ activation functions trained
with backpropagation through time. (ii) A long-short term memory (LSTM)
recurrent neural network \citep{hochreiterLongShortTermMemory1997}, also trained
with backpropagation through time. (iii) A gated recurrent unit (GRU) recurrent
neural network \cite{choPropertiesNeuralMachine2014}. (iv) A standard
encoder-only transformer neural network model \cite{vaswaniAttentionAllYou2017}.
(v) A logistic regression using a bag-of-words representation of each sentence
as input features. All the models are trained with batches of training data
using the Adam optimization algorithm \cite{kingmaAdamMethodStochastic2015}.
Each model's hyperparameters are chosen to ensure they have a similar number of
trainable parameters except for the logistic regression whose parameter count is
solely determined by the input and output dimensions.

\begin{minipage}[htbp]{.66\linewidth}
% \begin{table}[htbp]
 \vspace{21pt}
  \centering
  \begin{tabular}{c|cc}
\toprule
& \bfseries WADE $\times 10^{-2}$ (std.) $\uparrow$ & \bfseries Max test accuracy (std.) $\uparrow$ \\
\midrule
    RNN & 1.028 ${\scriptstyle \pm0.283}$ & 0.705 ${\scriptstyle \pm0.076 }$ \\
LSTM & 2.280 ${\scriptstyle \pm0.303}$ & 0.902 ${\scriptstyle \pm0.002 }$ \\
GRU & 2.711 ${\scriptstyle \pm0.318}$ & \bfseries 0.904 ${\scriptstyle \pm0.002 }$ \\
Linear & 3.737 ${\scriptstyle \pm0.745}$ & 0.862 ${\scriptstyle \pm0.000 }$ \\
Transformer & \bfseries 8.716 ${\scriptstyle \pm0.720}$ & 0.872 ${\scriptstyle \pm0.003 }$ \\

    \bottomrule
  \end{tabular}
  \captionof{table}{Comparison of our new WADE metric to assess learning speed and the standard maximum test accuracy on the IMDB
    classification dataset. Results are averaged over 50 separate runs. ($\uparrow$
    indicates that higher is better).
    \label{tab:comp-sup-acc}}
% \end{table}
\end{minipage}
\hfill
\begin{minipage}[htbp]{.32\linewidth}
  \vfill
  \centering
  \includegraphics[width=\linewidth]{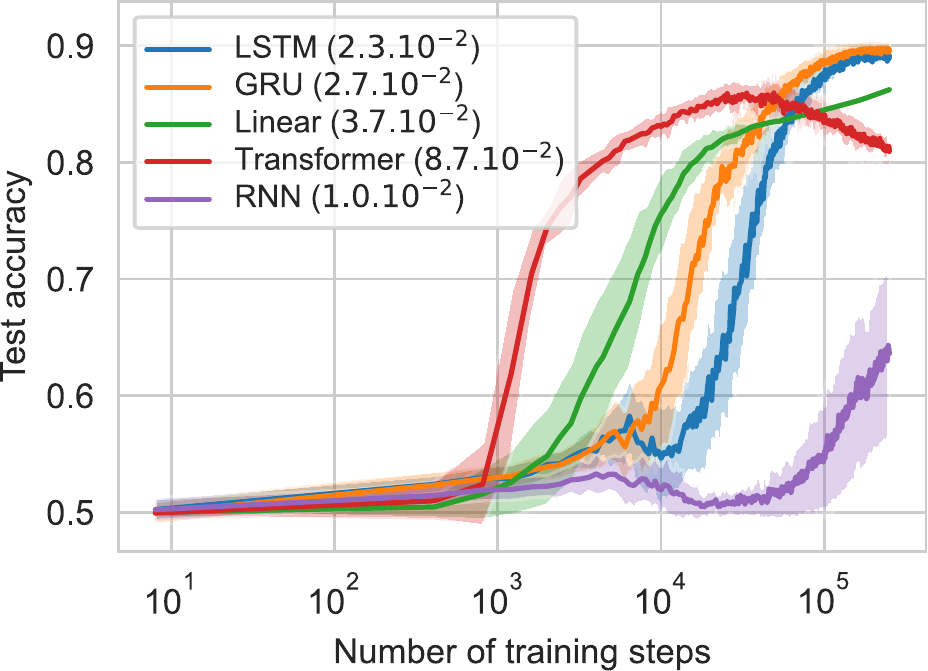}
  \vspace{-20pt}
  \captionof{figure}{Test accuracy curves for each model. Shaded areas are
    $\pm 1\sigma$ around the average over 50 runs.\label{fig:wade-sup}}%
  \vfill
\end{minipage}

Table~\Ref{tab:comp-sup-acc} shows the results of all models on the IMDB dataset
reporting both the standard test accuracy as well as our new WADE metric
measuring the learning speed of the different models. We report the maximum test
accuracy observed during training. This corresponds to the model checkpoint that
would be selected with a validation set before using it on test data. The
transformer model learns the fastest of all, as shown by its higher WADE score,
but it does not reach test accuracy values as high as the GRU and LSTM model. We
think that this could be attributed to the transformer model over-fitting on the
available data as discussed below.

According to the WADE metric, the RNN is the worst (i.e., slowest to learn) model
with a score of $1.04 \times 10^{-2}$ whereas the linear model and the transformer
are the fastest with a score above $8 \times 10^{-2}$. The transformer seems to have
more reliable learning speeds than the linear model as shown by the lower
standard deviation ($0.7\times 10 ^{-2}$ versus $4.3 \times 10^{-2}$). The LSTM and GRU
models have intermediate and stable learning speeds.
%You can clearly see on the graph why the GRU has slightly better score than the LSTM, with the former's curve being slightly above the latter's.
The GRU has slightly better WADE score than the LSTM, which can be also seen by inspecting the learning curves in figure~\ref{fig:wade-sup} where the GRU is consistently above the LSTM.

However, when we look at the maximum test accuracy we obtain a different ordering,
with LSTM and GRU performing best with slightly above $90\%$ accuracy followed by the
transformer and the linear model (around $86\%$) and finally the RNN with the lowest
accuracy. The LSTM, GRU and linear models all seem to have converged at the end
of the experiment. The transformer's lower final accuracy despite it being
considered a state of the art model may be explained by over-fitting (apparent
on the graph with the test accuracy score going down). Moreover, the RNN clearly
hasn't converged at the end of the experiment which also explains the relatively
low max accuracy score.

Our new WADE metric gives a complementary view of a model's abilities, which may be significantly different from what can be obtained from the usual
performance metrics focused on the final accuracy of the model.

\section{New benchmark: compared methods\label{sec:compared-methods}}

In the following sections, we carry out experiments to measure \acf{WADE} on
the benchmark described in Table~\ref{tab:all-tasks} and introduced in section~\ref{sec:descr-tasks-benchm}. We compare several
baseline models to understand their learning efficiency: three of the most common
sequential machine learning models for which all parameters are trained --- RNNs, LSTMs, and transformes --- with two methods using reservoir computing: the echo-state networks
(ESN, with a random RNN reservoir) and reservoir cellular automata (ReCA, with a
cellular automaton reservoir) for which only a fraction of the parameters are
learned. Table~\ref{tab:methods-summary} presents all the methods we study in
our experiments.
\begin{table}[h]
  \centering
  \begin{tabular}{lll}
    \toprule
    \multicolumn{2}{l}{\bfseries Methods} & \bfseries Type \\
    \midrule
   RNN & Recurrent neural networks  & \multirow{3}{*}{Fully trained}\\
   LSTM & Long-short term memory networks  & \\
   Transformer &  & \\
    \midrule
   ESN & Echo-state networks  & \multirow{2}{*}{Reservoir-based}\\
   ReCA & Reservoir Cellular Automata  & \\
    \bottomrule
  \end{tabular}
  \caption{Summary of the compared models: three are fully trained, and two are
    reservoir-based.}
\label{tab:methods-summary}
\end{table}

In all our tasks, the input data is assumed to be categorical and sequential.
Tokens are observed one by one in sequence, we define this input as
$X = [X_{1}, \ldots, X_{t}, \ldots],\; t\in \mathbb{N}, \; \forall t \ X_{t} \in \mathcal{X} \subset \mathbb{N}$, where $t$ is the
time index of the sequence, each $X_{i}$ is a token corresponding to time index
$i$, and $\mathcal{X}$ is the set of numbered input categories --- or different tokens in the
vocabulary. First, each categorical input vector is one-hot encoded into a
vector of size $L$, where $L = |\mathcal{X}|$ is the size of the input vocabulary. We define
$\vx_{t}$ as the encoded vector form of $X_{t}$, and we have
$\forall t,\; \vx_{t} \in {\{0, 1\}}^{L}$, with $\sum_{i = 1}^{L}{(\vx_{t})}_{i} = 1$.
Since the tasks are designed to be generally compatible with language modeling,
the input and output vocabularies are the same.

\subsection{Fully trained sequential models (RNN, LSTM and Transformer)}
We first study two standard supervised recurrent models: (i) an Elman recurrent
neural network (RNN) \citep{elmanFindingStructureTime1990} with $\tanh$
activation functions trained with backpropagation through time. (ii) A
long-short term memory (LSTM) recurrent neural network
\citep{hochreiterLongShortTermMemory1997}, also trained with backpropagation
through time. (iii) An encoder-only transformer model with positional encoding
\citep{vaswaniAttentionAllYou2017}.

The three models are trained with a batched Adam optimization algorithm
\citep{kingmaAdamMethodStochastic2015} to minimize a cross-entropy loss function
between the predicted and target tokens. No other training device, such as
dropout, regularization, or normalization, is used for these fully trained baselines.

\subsection{Echo-state networks and reservoir cellular automata\label{sec:echo-state-networks}}

We use %the notation and
names adapted from \citet{jaegerLongShortTermMemory2012} in
this section. An echo-state network is a random recurrent neural network with
frozen weights and skip-connections. Its random weights are sampled in a
specific way so that it performs random combinations of input vectors with its
current state while keeping a history of past inputs \citep{jaegerEchoStateApproach2001}.

% \subsection{Reservoir cellular automata}
Reservoir cellular automaton (ReCA) is a model similar to echo-state networks
% (described in Section~\ref{sec:echo-state-networks})
but where a cellular automaton (CA) replaces the random RNN. The cellular
automaton can be seen as a \ac{RNN} with additional weigth-sharing similar to
convolutional neural network. The special structure of a recurrent CA update
makes it more likely than random RNNs to generate complex structures
\citep{wolframStatisticalMechanicsCellular1983, wolframNewKindScience2002}.
% A CA can also be seen as a special RNN with a large amount of parameter sharing.
Because CAs were not designed to make use of inputs or produce outputs, we
extend the model to make it accept input vectors and to make reading from the CA
state possible (details in Section~\ref{sec:app-ca-res}).

\subsection{Experimental set-up}
We ran 100 separate experiments with different random seeds for each of the CA
rules, the RNN, LSTM, Transformer, and ESN on each task in the benchmark. These experiments
have a separate input projection matrix for the CA, different random weights
for the ESN, and different weight initialization for the supervised baselines.
The task inputs are also generated from a new seed for every experiment, but
we reuse the seeds for the same experiment on different models to ensure they
were trained with the same data. Intervals of one standard deviation for these multiple
experiments are reported in the result graphs.
%as shaded areas around the curve in Figure~\ref{fig:all_metrics}.
The code to reproduce our experiments is
available on GitHub\footnote{\url{https://github.com/hugcis/benchmark_learning_efficiency}}.

\subsubsection{Training parameters}
For each experiment, we generate 1200 random examples from the task generator.
We split this set randomly into a training set with 80\% of the data --- 960
examples --- and a test set with the remaining 240 examples. The reservoir is run
on each training example for the reservoir-based models, which creates the
input features for training the decoder. The sequential supervised models use
batches of single sequences with the Adam algorithm
\citep{kingmaAdamMethodStochastic2015}. They are trained for ten epochs in total. With
reservoir models, only the last linear layer (the decoder) is trained. We minimize the
cross-entropy loss with stochastic gradient descent (SGD), doing only a single
pass over the 960 training examples.

Every few training steps, we generate the output predictions on the testing set,
decode it, and compute the test accuracy for our WADE metric. Supervision is only
applied on tokens that can be predicted --- similar to masked language modeling,
\eg only the answer token is used in the symbol counting or question answering
tasks. Accuracy is also computed for these symbols only.

\section{Results}

We report the final weighted average data efficiency (WADE) scores on our
benchmark in Table \ref{tab:summary} and accuracy results for comparison in Table~\ref{tab:accuracy_all}. We include the best
elementary reservoir cellular automaton (ReCA) and echo
state network (ESN) with the same internal state size for each task.

\begin{figure}[htbp]
  \centering
  \includegraphics[width=.96\linewidth]{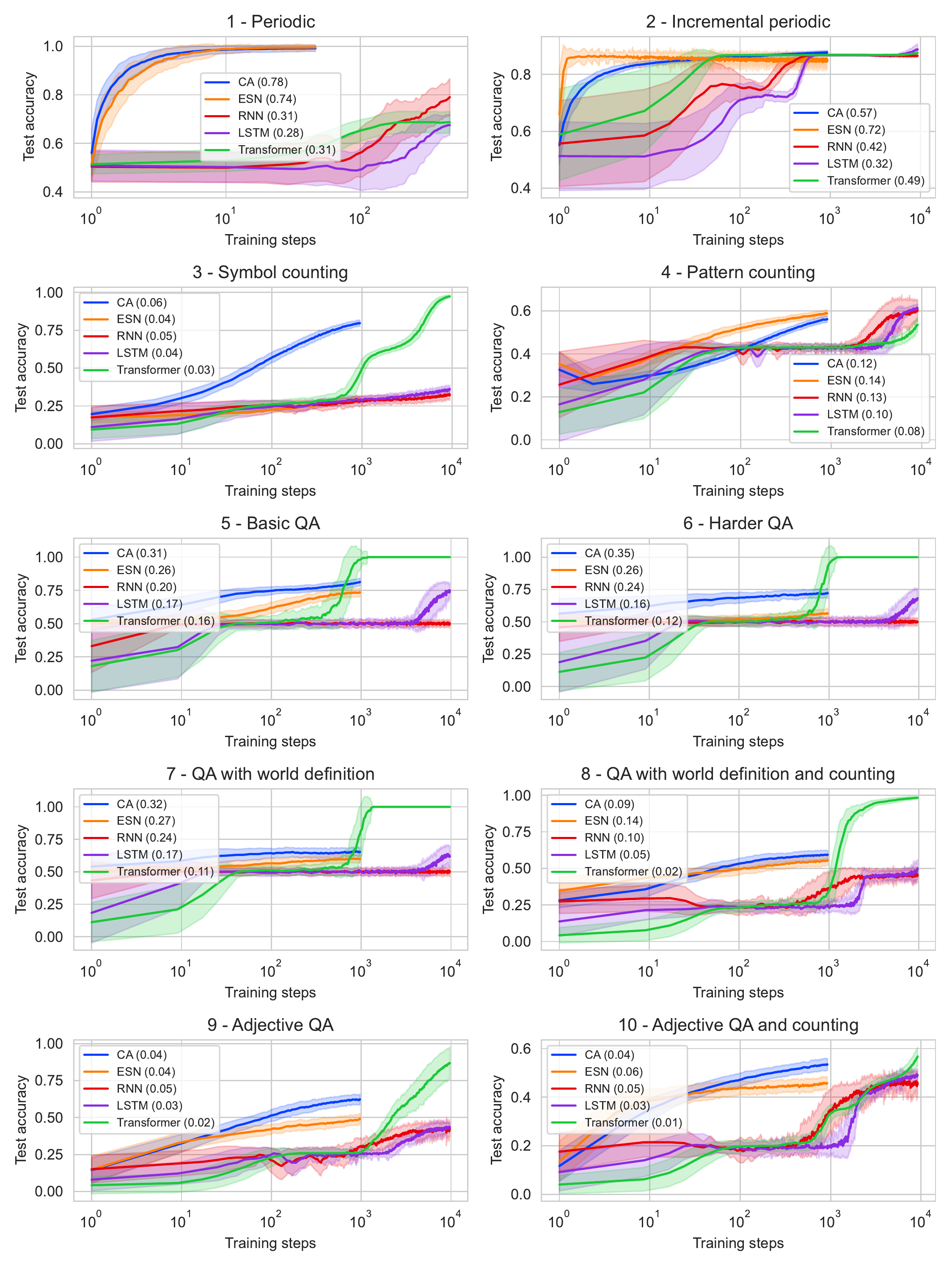}
  \caption{Average learning curves and WADE scores (shown in parentheses in the
    legend) for each task in the benchmark. The dark blue curves represent the
    best elementary cellular automata (CA) rule and the green curves represent
    the echo-state networks (ESN). Note that the x-axis is logarithmic, showing
    ten times more training steps for the RNN and LSTM. Shaded areas represent
    one standard deviation around the average over the 100 different
    experiments.}
  \label{fig:all_metrics}
\end{figure}

\begin{table}[htbp]
  \centering
    \begin{tabular}{lcc|ccc|c}
      \toprule
      \multirow{2}{*}{\bfseries Task ID - Name} & \multicolumn{2}{c|}{Reservoir} &
        \multicolumn{3}{c|}{Fully supervised} & \multirow{2}{*}{\bfseries \shortstack{Human \\ expert}}\\
& \bfseries ReCA & \bfseries ESN & \bfseries RNN & \bfseries LSTM & \bfseries Transformer &  \\
\midrule

\bfseries 1 - Periodic & \bfseries 0.78 ${\scriptscriptstyle \pm0.05 }$ &  0.74 ${\scriptscriptstyle \pm0.05 }$ &  0.31 ${\scriptscriptstyle \pm0.06 }$ &  0.28 ${\scriptscriptstyle \pm0.07 }$  &  0.31 ${\scriptscriptstyle \pm0.04 }$& 1.00 \\
\bfseries 2 - Incremental periodic &  0.57 ${\scriptscriptstyle \pm0.02 }$ & \bfseries 0.72 ${\scriptscriptstyle \pm0.02 }$ &  0.42 ${\scriptscriptstyle \pm0.16 }$ &  0.32 ${\scriptscriptstyle \pm0.11 }$  &  0.49 ${\scriptscriptstyle \pm0.15 }$& 1.00 \\
\bfseries 3 - Symbol counting & \bfseries 0.06 ${\scriptscriptstyle \pm0.01 }$ &  0.04 ${\scriptscriptstyle \pm0.01 }$ &  0.05 ${\scriptscriptstyle \pm0.02 }$ &  0.04 ${\scriptscriptstyle \pm0.02 }$  &  0.03 ${\scriptscriptstyle \pm0.01 }$& 0.11 \\
\bfseries 4 - Pattern counting &  0.12 ${\scriptscriptstyle \pm0.04 }$ & \bfseries 0.14 ${\scriptscriptstyle \pm0.03 }$ &  0.13 ${\scriptscriptstyle \pm0.04 }$ &  0.10 ${\scriptscriptstyle \pm0.05 }$  &  0.08 ${\scriptscriptstyle \pm0.03 }$& 0.09 \\
\bfseries 5 - Basic question answering (QA) & \bfseries 0.31 ${\scriptscriptstyle \pm0.08 }$ &  0.26 ${\scriptscriptstyle \pm0.03 }$ &  0.20 ${\scriptscriptstyle \pm0.06 }$ &  0.17 ${\scriptscriptstyle \pm0.08 }$  &  0.16 ${\scriptscriptstyle \pm0.07 }$& 0.15 \\
\bfseries 6 - Harder QA & \bfseries 0.35 ${\scriptscriptstyle \pm0.11 }$ &  0.26 ${\scriptscriptstyle \pm0.03 }$ &  0.24 ${\scriptscriptstyle \pm0.05 }$ &  0.16 ${\scriptscriptstyle \pm0.07 }$  &  0.12 ${\scriptscriptstyle \pm0.06 }$& 0.12 \\
\bfseries 7 - QA with world def. & \bfseries 0.32 ${\scriptscriptstyle \pm0.10 }$ &  0.27 ${\scriptscriptstyle \pm0.04 }$ &  0.24 ${\scriptscriptstyle \pm0.05 }$ &  0.17 ${\scriptscriptstyle \pm0.07 }$  &  0.11 ${\scriptscriptstyle \pm0.06 }$& --- \\
\bfseries 8 - QA with world def. \& counting &  0.09 ${\scriptscriptstyle \pm0.03 }$ & \bfseries 0.14 ${\scriptscriptstyle \pm0.06 }$ &  0.10 ${\scriptscriptstyle \pm0.05 }$ &  0.05 ${\scriptscriptstyle \pm0.02 }$  &  0.02 ${\scriptscriptstyle \pm0.01 }$& --- \\
\bfseries 9 - Adjective QA &  0.04 ${\scriptscriptstyle \pm0.03 }$ &  0.04 ${\scriptscriptstyle \pm0.04 }$ & \bfseries 0.05 ${\scriptscriptstyle \pm0.03 }$ &  0.03 ${\scriptscriptstyle \pm0.02 }$  &  0.02 ${\scriptscriptstyle \pm0.01 }$& --- \\
\bfseries 10 - Adjective QA \& counting &  0.04 ${\scriptscriptstyle \pm0.02 }$ & \bfseries 0.06 ${\scriptscriptstyle \pm0.02 }$ &  0.05 ${\scriptscriptstyle \pm0.02 }$ &  0.03 ${\scriptscriptstyle \pm0.02 }$  &  0.01 ${\scriptscriptstyle \pm0.01 }$& --- \\
      \bottomrule
    \end{tabular}
    \caption{Comparison of WADE scores (also shown in parentheses in the legend
      in figure~\ref{fig:all_metrics}, higher is better) of the best cellular
      automaton rules (\textbf{ReCA}, \citet{yilmazReservoirComputingUsing2014}) for each task against an echo-state
      network (\textbf{ESN}, \citet{jaegerEchoStateApproach2001}), a \textbf{RNN}, a \textbf{LSTM} \citep{hochreiterLongShortTermMemory1997}, and a
      \textbf{Transformer} \citep{vaswaniAttentionAllYou2017} with the same number of parameters.
      %as the cellular automaton reservoir.
      The dash ``---'' indicates that the task was too
      difficult to complete from memory alone for the human
      expert. Accuracy scores are also reported in Table \ref{tab:accuracy_all}.}\label{tab:summary}
\end{table}
Figure \ref{fig:all_metrics} shows the learning curves for the best reservoir cellular
automaton, as well as the echo-state network, recurrent neural network (RNNs), LSTM, and Transformer
on all ten tasks. Interestingly, the reservoir based-models are consistently more efficient
learners than the fully supervised methods, reaching better accuracy in much
fewer training steps. For example, for tasks 5, 6, and 7, the LSTM needs ten
times more steps to approach the accuracy the reservoir CA model reached in less
than 1000 training steps.

Echo-state networks (ESN) and reservoir cellular automata (ReCA) appear to learn at similar
rates, with no clear advantage for one or the other as each model outperforms
the other on five tasks out of ten. On tasks 2, 4, 8, and 10 the ESN is visibly
faster than the ReCA (see curves Figure \ref{fig:all_metrics}), which explains
the ESN's higher WADE scores even though ReCA reaches higher accuracy values
after several more training steps.

\begin{table}[htbp]
  \centering
    \begin{tabular}{lcc|ccc}
      \toprule
      \multirow{2}{*}{\bfseries Task ID - Name} & \multicolumn{2}{c|}{Reservoir} &
        \multicolumn{3}{c}{Fully supervised} \\
& \bfseries ReCA & \bfseries ESN & \bfseries RNN & \bfseries LSTM & \bfseries Transformer  \\
\midrule
\bfseries 1 - Periodic & 0.99 ${\scriptscriptstyle \pm 0.01}$ & \bfseries 1.00 ${\scriptscriptstyle \pm 0.01}$ & 0.79 ${\scriptscriptstyle \pm 0.08}$ & 0.68 ${\scriptscriptstyle \pm 0.04}$ & 0.69 ${\scriptscriptstyle \pm 0.04}$  \\
\bfseries 2 - Incremental periodic & 0.88 ${\scriptscriptstyle \pm 0.01}$ & 0.87 ${\scriptscriptstyle \pm 0.03}$ & 0.87 ${\scriptscriptstyle \pm 0.00}$ & \bfseries 0.89 ${\scriptscriptstyle \pm 0.02}$ & 0.88 ${\scriptscriptstyle \pm 0.00}$  \\
\bfseries 3 - Symbol counting & 0.80 ${\scriptscriptstyle \pm 0.02}$ & 0.30 ${\scriptscriptstyle \pm 0.02}$ & 0.33 ${\scriptscriptstyle \pm 0.03}$ & 0.36 ${\scriptscriptstyle \pm 0.03}$ & \bfseries 0.97 ${\scriptscriptstyle \pm 0.01}$  \\
\bfseries 4 - Pattern counting & 0.56 ${\scriptscriptstyle \pm 0.01}$ & 0.59 ${\scriptscriptstyle \pm 0.02}$ & 0.61 ${\scriptscriptstyle \pm 0.05}$ & \bfseries 0.61 ${\scriptscriptstyle \pm 0.02}$ & 0.54 ${\scriptscriptstyle \pm 0.03}$  \\
\bfseries 5 - Basic QA & 0.81 ${\scriptscriptstyle \pm 0.03}$ & 0.73 ${\scriptscriptstyle \pm 0.03}$ & 0.51 ${\scriptscriptstyle \pm 0.02}$ & 0.75 ${\scriptscriptstyle \pm 0.05}$ & \bfseries 1.00 ${\scriptscriptstyle \pm 0.00}$  \\
\bfseries 6 - Harder QA & 0.72 ${\scriptscriptstyle \pm 0.04}$ & 0.57 ${\scriptscriptstyle \pm 0.03}$ & 0.51 ${\scriptscriptstyle \pm 0.03}$ & 0.68 ${\scriptscriptstyle \pm 0.07}$ & \bfseries 1.00 ${\scriptscriptstyle \pm 0.00}$  \\
\bfseries 7 - QA with world def. & 0.66 ${\scriptscriptstyle \pm 0.03}$ & 0.60 ${\scriptscriptstyle \pm 0.03}$ & 0.51 ${\scriptscriptstyle \pm 0.04}$ & 0.64 ${\scriptscriptstyle \pm 0.07}$ & \bfseries 1.00 ${\scriptscriptstyle \pm 0.00}$  \\
\bfseries 8 - QA with world def. \& counting & 0.59 ${\scriptscriptstyle \pm 0.03}$ & 0.56 ${\scriptscriptstyle \pm 0.03}$ & 0.47 ${\scriptscriptstyle \pm 0.05}$ & 0.50 ${\scriptscriptstyle \pm 0.06}$ & \bfseries 0.98 ${\scriptscriptstyle \pm 0.01}$  \\
\bfseries 9 - Adjective QA & 0.62 ${\scriptscriptstyle \pm 0.04}$ & 0.49 ${\scriptscriptstyle \pm 0.03}$ & 0.43 ${\scriptscriptstyle \pm 0.04}$ & 0.44 ${\scriptscriptstyle \pm 0.03}$ & \bfseries 0.87 ${\scriptscriptstyle \pm 0.11}$  \\
\bfseries 10 - Adjective QA \& counting & 0.54 ${\scriptscriptstyle \pm 0.02}$ & 0.46 ${\scriptscriptstyle \pm 0.03}$ & 0.47 ${\scriptscriptstyle \pm 0.04}$ & 0.49 ${\scriptscriptstyle \pm 0.03}$ & \bfseries 0.57 ${\scriptscriptstyle \pm 0.04}$  \\

      \bottomrule
    \end{tabular}
    \caption{Comparison of accuracy scores (higher is better) of the
      \textbf{ReCA}, \textbf{ESN}, \textbf{RNN}, \textbf{LSTM}, and
      \textbf{Transformer} models with a similar number of parameters. In contrast to the
      results of table~\ref{tab:summary}, the fully-supervised models are more
      performant when we measure the accuracy, as shown here, but do not necessarily do as well when we measure  the speed of learning, as shown in table~\ref{tab:summary}.
    }\label{tab:accuracy_all}
\end{table}

Even if they lack slightly in accuracy, as seen for example on the curves of task 4,
the two reservoir-based methods consistently outperform the fully supervised
sequential methods (RNNs and LSTMs) in terms of learning speeds. This better
learning efficiency could be explained by the internal state structure
introduced by the CA rules and the special form of the ESN random matrix which
favors memory retention whereas usual recurrent network initialization does not
--- we initialize the weights of the full trained models from a uniform
distribution $\mathcal{U}( - \sqrt{h^{-1}} ,\sqrt{h^{-1}})$ in our experiments, where $h$
is the hidden size.

The human expert scores are estimated based on the authors' performance on
obfuscated versions of the tasks --- all input tokens are mapped to random
symbols. These scores may appear surprisingly low in some of these experiments.
Although the task examples presented in Section~\ref{sec:descr-tasks-benchm}
seem easy to understand, we rely heavily on our prior knowledge about the
symbols to understand the patterns --- the words in tasks 5--10 or the numbers in
tasks 3 and 4. These symbols are randomly remapped in our human experiments, and
more examples as well as a good memory are needed to understand the tasks and
learn the mapping itself, hence the slower learning.

It is interesting to note that RNNs also seem to perform better than LSTMs and
Transformers at the beginning of the training in all tasks but the first two.
Even though the Transformer is nowadays generally considered a superior model,
vanilla RNNs may still remain competitive in the very low data and computation
regimes. This is especially pronounced for tasks 8, 9, and 10, where the RNN
test accuracy curve starts increasing significantly 1000 steps before the LSTM.
After several epochs, the Transformer still outperforms other alternatives in
terms of test accuracy on most tasks.
%limitations

We note that we implicitly assume a fixed cost per training iteration in our
experiments and that each training example is seen once. Without these
requirements, one could achieve higher WADE results at the cost of additional
computations and memory usage by using \emph{replay}-inspired methods
\citep{hintonUsingFastWeights1987, robinsCatastrophicForgettingNeural1993a,
  gepperthBioInspiredIncrementalLearning2016,
  rebuffiIcarlIncrementalClassifier2017} that retrain the models with past
stored inputs. The results in Table~\ref{tab:summary} used each input sequence
once.

\section{Conclusions}

Learning speed and data efficiency are essential components of any learning
system. Our learning speed metric can offer a novel perspective on several
machine learning models.
%old and new machine learning models.
Instead of focusing on pure performance, measuring and comparing the data
efficiency of different models will hopefully lead to better systems for online
and continual learning.

%that can be used in online and continual learning contexts.

Even with the right metric, it is difficult to thoroughly evaluate a model's
ability to learn efficiently. Our benchmark evaluates a range of problem
difficulties which would be challenging to construct by combining or
manipulating existing datasets.
%by using multiple other existing datasets or manipulating a single dataset.
However, the tasks remain easy to understand and use. Since the tasks are
language-based, they can be further mixed or chained to create continual
learning problems and may also be easily extended in a follow-up work.

%Take-home message
We study lesser-known machine learning models based on evolving states of
complex systems that can learn through self-organization.
A complex dynamical system comprises many interacting agents and evolves over
time according to a fixed update rule. These agents can be hidden neurons of a
\ac{RNN} or nodes in a graph. Such systems often exhibit emergent global
dynamics resulting from the actions of its parts rather than the decisions of a
central controller. These dynamics lead to surprisingly complex behavior, which
can be random, chaotic, or may lead to unbounded growth of complexity
\citep{boccaraModelingComplexSystems2010}.
Due to these properties, %this class of systems
systems based on reservoir computing may be a promising alternative addressing
the shortcomings of standard supervised models.

Surprisingly, %complex systems-based models
such models achieve remarkable learning efficiency compared to the more standard
sequential supervised models trained with stochastic gradient-based learning.
%stochastic gradient methods.
The complex systems-based models consistently outperform sequential supervised
methods and even achieve better learning efficiency than humans on some tasks.
They demonstrate more efficient learning on our benchmark at a fraction of the
computational and data cost of the conventional models. We believe that more
advanced models of this type could lead to more robust and data efficient
machine learning in the future, especially in low-data applications or problems
where supervision is limited. Complex systems are underexplored and seem worth
investigating further for building the next generation of learning
algorithms. % that learn.

\section*{Acknowledgments}
This work was partly supported by the European Regional Development Fund under
the project IMPACT (reg. no. CZ.02.1.01/0.0/0.0/15 003/0000468).
% Interesting results with the complex systems on our tasks

\bibliography{library}
\bibliographystyle{collas2022_conference}

\newpage 

\appendix

\section*{Appendix}

In this appendix we provide details of the human evaluation
(Section~\ref{sec:human}), we list all the experimental parameters used in
the evaluation (Section~\ref{sec:parameters}), and describe the reservoir models
in more details (Section~\ref{sec:res-models}).

\section{Human performance evaluation}
\label{sec:human}
The proposed benchmark is relatively simple to understand and hence one may ask what would be the human performance and what WADE values this would correspond to. In
particular, the language tasks appear to be readily solvable just from one's
understanding of the English language. However, similar to how words are encoded
as vectors before being processed by a neural network, we need to remove any
language-based prior to make the comparison with human performance fair. This
could be achieved in a number of ways. For example, if we map all the available
tokens to random letters of the alphabet the apparently trivial task
{\small \begin{align*}
          & \texttt{I DO NOT SEE PAUL . DO I SEE PAUL ? NO} \\
          & \texttt{I HEAR JAMES BUT I DO NOT HEAR PAUL AND JOHN . DO I HEAR JAMES ? YES}
\end{align*}}

would become significantly less obvious to humans
{\small
\begin{align*}
  & \texttt{w K k A D r K w A D l W} \\
  & \texttt{w Z t g w K k Z D G C r K w Z t l H}.
\end{align*}
}

One would need to read through several of these sentences to identify patterns
such as the fact that \texttt{W} and \texttt{H} represent Yes or No or that
\texttt{r} plays the role of a full stop.

We apply the following procedure in our human evaluation experiments:
\begin{enumerate}
  \item Choose a random task among 10.
  \item Apply a random mapping from the task token to letters of the alphabet.
  \item Present the sequences one by one with the tokens to be predicted hidden.
  \item The user enters his predicted answer.
  \item The valid sequence is shown so the user can learn from their mistakes.
  \item Step 3-5 are repeated until 10 sequences are correctly solved in a row.
  Then we go back to step 2 with a new task.
\end{enumerate}
We add the requirement that no external device such as pen and paper or
note-taking program should be used during the experiment so the user relies
solely on their memory. The accuracy is computed by counting the number of right
answers in a row. For example, three right answers in a row would correspond to an
accuracy of 30\%, while ten right answers in a row is 100\% which also
corresponds to a switch of a task. This accuracy score is attributed to the first
correct answer of the series of correct answers. This means that a series of five correct answers starting from question 6 until question 11 will correspond to reaching the accuracy of 50\% on question 6.
This ensures that ten correct answers on the
first attempt will yield a WADE score of 1.

\section{Experimental parameters}
\label{sec:parameters}
We report all the parameters used in our experiments in table \ref{tab:appendix-params}:
\begin{table}[htbp]
  \centering
  \begin{tabular}{lp{.2\linewidth}p{.35\linewidth}}
    \toprule
    & \bfseries Reservoir-based & \bfseries Fully trained \\
    \midrule
    \bfseries Total number of sequences per task & 1200 & 1200 \\
    \bfseries Number of training sequences & 960 & 960 \\
    \bfseries Number of testing sequences & 240 & 240 \\
    \bfseries Passes over the data (epochs) & 1 & 10 \\
    \bfseries Random runs per task & 100 & 100\\
    \bfseries Algorithm & SGD & Adam \citep{kingmaAdamMethodStochastic2015}\\
    \bfseries Learning rate & $0.001$ & $0.001$ \\
    \bfseries Regularization & weight decay $0.001$ & None \\
    \bfseries Internal state (hidden) size & $1800$ & $h$ task-dependent \\
    \bfseries Number of trainable parameters & $1800 \times \text{dictionary\_size}$ &
      $h^{2} + 2 \times h \times \text{dictionary\_size}$ for the RNN\textsuperscript{2}\\
    \bfseries Output non-linearity & Softmax & Softmax \\
    \bfseries Internal non-linearity & Not applicable & $\tanh$ \\
    \bfseries Batch size  & 1 & 1 \\
    \bottomrule
  \end{tabular}
  \caption{Experimental parameters common to all tasks. Some of the sizes (such
    as $h$) vary from task to task.}
  \label{tab:appendix-params}
\end{table}

\subsection{Task generation parameters\label{sec:task-gen-params}}
The tasks are generated according to the procedures described in
Section~\ref{sec:tasks}. They have some parameters that allow the
difficulty of the task to be adjusted. We report results on ten tasks. The parameters
chosen for our experiments are listed below:
\vspace{-8pt}
\begin{description}
  \item[Periodic (1) and Increasing period (2):] Sequences are generated from patterns of
        length between 1 and 10.
  \item[Easy symbol counting (3):] Available symbols are \texttt{A}, \texttt{B} and
        \texttt{C}. Generated sequences have between 1 and 10 of these symbols
        in the prompt, and the query has either one, two, or all three symbols.
  \item[Hard symbol counting (4):] Available symbols are the same as for the previous
        task. Generated sequences have between 1 and 45 symbols
        with separators in the prompt. There is a minimum of 1 query, and can be
        as many queries as there are different patterns in the prompt.
  \item[Question answering (5):] There are five available names and two verbs. The
        prompt consists of between one and five names, and the question is about one
        of these names.
  \item[Harder question answering :] There are eleven available names and five
        verbs. The prompt is made of between one and five names, and the
        question is about one of these names.
  \item[Question answering with world (7), with counting (8):] There are thirteen available
        names and seven verbs.
  \item[Adjective question answering (9), with counting (10):] There are eight
        available names, six verbs, four color adjectives, and five size
        adjectives. The prompt consists of between one and six statements, and
        there are up to eight questions.
\end{description}

\vspace{-8pt}
We chose relatively small values of these parameters because we are interested in simple tasks
so that models that can learn from less than 100 examples. Moreover, the
number of possible sentences generated from these small values is already huge,
with more than $10^{35}$ possible sentences for task 4, for example. These
parameters could also be varied dynamically to change the task's difficulty when
needed or turn each task into multiple sub-tasks or levels of difficulty. We
chose the ten separate tasks/settings pairs listed above to get a broad overview of our
benchmarked models on a range of task difficulties.
\footnotetext{$h$ is chosen to match the number of parameters of the reservoir-based methods.}

\section{Reservoir models\label{sec:res-models}}
We describe in more details the two reservoir-based models used in our experiments.
\subsection{Echo-state network}
The echo-state network is based on the
following update equation:
\begin{equation}
  \label{eq:esn}
  \vr_{t + 1} = (1 - \alpha) \vr_{t} +
  \beta \tanh(\mW\: \vr_{t} + \mW_{\text{in}}\: \vx_{t + 1}),
\end{equation}
where $\vr_{t}$ is the $K$-dimensional state vector -- corresponding to the hidden neurons
--- at time $t$, $\beta$ is the leaking rate, $\mW \in \mathbb{R}^{K \times K}$ is a sparsely
connected random hidden layer matrix, and $\mW_{\text{in}} \in \mathbb{R}^{L \times K}$ is the
input projection matrix. The matrices $\mW$ and $\mW_{\text{in}}$ are
initialized randomly using the recommendations of
\citet{jaegerLongShortTermMemory2012}: $\mW$ has an average of 10 non-zeros
entries per row, all sampled uniformly in $[-1, 1]$, $\mW_{\text{in}}$ has its
entries uniformly sampled in $[-1, 1]$.

The $L$-dimensional outputs are computed at times $t > 0$ as
\begin{equation}
  \label{eq:esn-res}
\tilde{\vx}_{t+ 1} = D(\vr_{t}),
\end{equation}
where $D: \mathbb{R}^{K} \rightarrow \mathbb{R}^{L}$ is a (trained) decoding function. For echo-state networks, the
decoder is often a linear transformation
$D(\vr_{t}) = \mW_{\text{out}} \vr_{t}$ where $\mW_{\text{out}}$ is a
$K \times L$-dimensional matrix. In our experiments, we set $\beta = 0$, which was
empirically observed to yield the best results on our tasks. The parameter $\beta$,
as well as the randomly sampled weight matrix, are sometimes tuned for each task
\citep{jaegerLongShortTermMemory2012}. We only use
default values to get a task-independent setup with
the least possible assumptions and to make the methods comparable.

\subsection{Reservoir cellular automata\label{sec:app-ca-res}}

The Cellular Automaton (CA) update function $\Phi$ is applied at time $t$ on the previous state vector
$\vs_{t - 1}$ --- also called the grid. The input vector $\vx_{t}$ at time $t$ is
projected onto a vector $\vp_{t}$ the same size as $\vs_{t}$ using a method proposed in
\cite{yilmazReservoirComputingUsing2014}. To combine inputs with CA states,
we XOR the two vectors together
\citep{gloverDynamicalLandscapeReservoir2021},
\begin{equation}
  \label{eq:ca-xor}
  \vs'_{t} = \vp_{t}\otimes \vs_{t},
\end{equation}
where $\vs'_{t}$ is the temporary CA state resulting from the combination of the
CA state and input vectors at time $t$. We then compute the next CA states by
applying the update function $\Phi$,
\begin{equation}
  \label{eq:ca-state}
  \vs_{t + 1} = \Phi(\vs'_{t}).
\end{equation}
The final reservoir state $\vr_{t + 1}$ is obtained by concatenating $r$ consecutive CA
states obtained from a single combined state $\vs'$ by applying $\Phi$ again. If
the size of the state is $n$, the resulting reservoir vector $\vr$ has a dimension
$K = r \times n$, where $r$ is defined above. As for the echo-state network (ESN), the $L$-dimensional output tokens are
calculated at times $t > 0$ as $\tilde{\vx}_{t+ 1} = D(\vr_{t})$ where
$D: \mathbb{R}^{K} \rightarrow \mathbb{R}^{L}$ is the (trained) decoding function. We use linear decoding functions in
our experiments such that $D(\vr_{t}) = \mW_{\text{out}}\vr_{t}$.

\end{document}